\documentclass[sigplan, camera-ready]{acmart}
\usepackage{booktabs} 

\setcopyright{none}

\acmDOI{}

\acmISBN{}

\acmConference[ASAIL'2017]{2nd Workshop on Automated Semantic
Analysis of Information in Legal Texts}{June 2017}{London, United Kingdom} 
\copyrightyear{}

\acmPrice{}

\usepackage{url}
\usepackage{latexsym}
\usepackage{color}

\begin{document}

\title{Exploring the Use of Text Classification in the Legal Domain}

\author{Octavia-Maria \c{S}ulea, Marcos Zampieri, Shervin Malmasi, Mihaela Vela, Liviu P. Dinu, Josef van Genabith}

\affiliation{\textsuperscript{1,5}University of Bucharest, Romania\\
\textsuperscript{2}University of Wolverhampton, United Kingdom\\ 
\textsuperscript{3}Harvard Medical School, United States\\ 
\textsuperscript{4,6}Saarland University, Germany\\
\textsuperscript{6}German Research Center for Artificial Intelligence (DFKI), Germany}

\renewcommand{\shortauthors}{\c{S}ulea et al.}


\begin{abstract}
In this paper, we investigate the application of text classification methods to support law professionals. We present several experiments applying machine learning techniques to predict with high accuracy the ruling of the French Supreme Court and the law area to which a case belongs to. We also investigate the influence of the time period in which a ruling was made on the form of the case description and the extent to which we need to mask information in a full case ruling to automatically obtain training and test data that resembles case descriptions. We developed a mean probability ensemble system combining the output of multiple SVM classifiers. We report results of 98\% average F1 score in predicting a case ruling, 96\% F1 score for predicting the law area of a case, and 87.07\% F1 score on estimating the date of a ruling.
\end{abstract}

\maketitle

\section{Introduction}



Text classification methods have been successfully applied to a number of NLP tasks and applications ranging from plagiarism \cite{barron2013plagiarism} and pastiche detection \cite{dinu2012pastiche} to estimating the period in which a text was published \cite{niculae2014temporal}. In this paper we discuss the application of text classification methods in the legal domain which, to the best of our knowledge, is relatively under-explored and to date its application has been mostly restricted to forensics \cite{de2001mining}. 

In this paper we argue that law professionals would greatly benefit from the type of automation provided by machine learning. This is particularly the case of legal research, more specifically the preparation a legal practitioner has to undertake before initiating or defending a case. The objective of the research reported in this paper is the following: given a case, law professionals have to make complex decisions including which area of law applies to a given case, what the ruling might be, which laws apply to the case, etc. Given the data available on previous court rulings, is it possible train text classification systems that are able to predict some of these decisions, given a textual ``draft'' case description provided by the professional? Such a system could act as a decision support system or at least a sanity check for law professionals. 

At present, law professionals have access to court ruling data through search portals\footnote{An example of a German website: \url{https://www.juris.de/jportal/index.jsp}} and keyword based search. In our work we want to go beyond this: instead of keyword based search, we use the full ``draft'' case description and text classification methods. For this purpose we acquire a large corpus of French court rulings with over 126,000 documents, spanning from the 1800s until the present day. We explore the use of lexical features and Support Vector Machine (SVM) ensembles on predicting the law area, the ruling, and on estimating the date of the ruling. We compare the results of our method to those reported by a previous study \cite{sulea2017predicting} which used the same data. Finally, we also investigate how much of the final case description attached to the judge's ruling needs to be masked to obtain a \emph{synthetic draft description}, close to what a lawyer would have at their disposal and how predictable the ruling is based on this description. All results reported in this paper are in fact on predictions based on these synthetic draft case descriptions, where what is to be predicted is masked in the training and test data and its descriptions in terms of features.


\section{Related Work}

While text classification methods were investigated and applied with commercial or forensic goals in mind for other areas (e.g. serving better content or products to users through user profiling \cite{consumer} and sentiment analysis, identifying potential criminals \cite{dark-triad}, crimes \cite{rada15}, or anti-social behavior \cite{danescu}), an area where these methods have been under-explored, although both commercial and forensic interests exist, is the legal domain.

Assuming that argumentation plays an important role in law practice,~\cite{Palau:2009} investigate to which extent one can automatically identify argumentative propositions in legal text, their argumentative function and structure. They use a corpus containing legal texts extracted from the European Court of Human Rights (ECHR) and classify argumentative vs. non-argumentative sentences with an accuracy of 80\%. 

Based on the association between a legal text and its domain label in a database of legal texts, \cite{Boella:2011} present a classification approach to identify the relevant domain to which a specific legal text belongs. Using TF-IDF weighting and Information Gain for feature selection and SVM for classification, ~\cite{Boella:2011} attain an f1-measure of 76\% for the identification of the domains related to a legal text and 97.5\% for the correct classification of a text into a specific domain.

Following the observation of a thematic structure in Canadian court rulings, where the intro, context, reasoning, and conclusion were found to be independent of the ruling itself, 
\cite{farzindar2004legal} present an automatic summarization of court rulings. \cite{galgani12} introduce a hybrid summarization system for legal text which combines hand crafted knowledge base rules with already existing automatic summarization techniques.

\cite{hachey2006extractive} proposed a system of classifying sentences for the task of summarizing court rulings and, with the use of SVM and Naive Bayes applied to Bag of Words, TF-IDF, and dense features (e.g. position of sentence in document), obtained 65\% f1 on 7 classes. Similarly, another study \cite{gonccalves2005evaluating} used BOW, POS tags, and TF-IDF to classify legal text in 3000 categories, based on a taxonomy of legal concepts, and reported 64\% and 79\% f1. 

For court ruling prediction, the task closest to our present work, a few papers have been published: \cite{Katz14}, using extremely randomized trees, reported 70\% accuracy in predicting the US Supreme Court's behavior and, more recently, \cite{wongchaisuwat2016} tackled the task of predicting patent litigation and time to litigation (TTL) and obtained lower than baseline 19\% f1 for predicting the litigation outcome, but a remarkable 87\% f1 for TTL prediction, when the interval considered was less than 4 years, and only 43\% f1 when the interval considered was narrowed down to less than a year. Among the most recent studies, \cite{aletras2016predicting} proposed a computational method to predict decisions of the European Court of Human Rights (ECRH) and \cite{sulea2017predicting} applied linear SVM classifiers to predict the decisions of the French Supreme Court using the same dataset presented in this paper. 

As evidenced in this section predicting court rulings is a new area for text classification methods and our paper contributes in this direction, achieving performance substantially higher than in previous work \cite{sulea2017predicting}.

\section{Corpus and Data Preparation}

In this paper, we use the diachronic collection of court rulings from the French Supreme Court, Court of Cassation ({\em Court de Cassation}). The complete collection\footnote{Acquired from \url{https://www.legifrance.gouv.fr}} contains 131,830 documents each consisting of a unique court ruling including metadata formatted in XML. Common metadata available in most documents include: law area, time stamp, case ruling (e.g. \emph{cassation},  \emph{rejet}, \emph{non-lieu}, etc.), case description, and cited laws. 
We use the metadata provided as "natural" labels to be predicted by the machine learning system. In order to simulate realistic test scenarios we automatically remove all mentions from the training and test data that explicitly refer to our target prediction classes. 


During pre-processing, we removed all duplicate and incomplete entries in the dataset. This resulted in a corpus comprising of 126,865 unique court rulings. Each instance contains a case description and four different types of labels: a law area, the date of ruling, the case ruling itself, and a list of articles and laws cited within the description. 



Taking the results by \cite{sulea2017predicting}, henceforth \c{S}ulea et al. (2017), as a baseline, in this paper we tackle 3 tasks:

\begin{enumerate}
\item Predicting the law area of cases and rulings (Section \ref{sec:area}).
\item Predicting the court ruling based on the case description (Section \ref{sec:ruling}).
\item Estimating the time span when a case description and a ruling were issued (Section \ref{sec:temporal}).
\end{enumerate}




\noindent Deciding which labels to use in each experiment was not trivial as this information was very often not explicit in the instances of the dataset and the distribution of instances in the classes was very unbalanced and sometimes inconsistent. For this reason, here we follow the decisions taken by  \c{S}ulea et al. (2017) and summarize them next. 

For task 1, predicting the law area of cases and rulings, out of 17 initial unique labels, the 8 labels that appeared in the corpus more than 200 times were kept. Table ~\ref{tab:1} shows the distribution of cases among each label. 

\vspace{2mm}

\begin{table}[h]
\begin{center}
\begin{tabular}{lr}
\hline 
\bf Law Area & \bf \# of cases\\ \hline

CHAMBRE\_SOCIALE 	& 33,139\\
CHAMBRE\_CIVILE\_1 	& 20,838 \\
CHAMBRE\_CIVILE\_2 	& 19,772\\
CHAMBRE\_CRIMINELLE & 18,476\\
CHAMBRE\_COMMERCIALE & 18,339\\
CHAMBRE\_CIVILE\_3 	& 15,095\\
ASSEMBLEE\_PLENIERE & 544\\
CHAMBRE\_MIXTE 		& 222\\
\hline
\end{tabular}
\end{center}
\caption{\label{tab:1} Distribution of cases according to the law area.}
\end{table}


\noindent For task 2, ruling prediction, we carry out two sets of experiments. A first set of experiments (6-class setup) considers only the first word within each label and only those labels which appeared more than 200 times in the corpus. This lead to an initial set of 6 unique labels: \emph{cassation}, \emph{annulation}, \emph{irrecevabilite}, \emph{rejet}, \emph{non-lieu}, and \emph{qpc} (\emph{question prioritaire de constitutionnalité}). In the second set of ruling prediction experiments (8-class setup), we consider all labels which had over 200 dataset entries and this time we did not reduce them to their first word as shown in Table ~\ref{tab:2}.

\vspace{4mm}

\begin{table}[h]
\begin{center}
\begin{tabular}{lr}
\hline 
\bf First-word ruling (6-class setup) & \bf \# of cases\\ \hline
rejet & 68,516 \\
cassation & 53,813 \\
irrecevabilite & 2,737 \\
qpc & 409 \\
annulation & 377 \\
non-lieu & 246 \\
\hline
\bf Full ruling (8-class setup) & \bf \# of cases\\ \hline

cassation & 37,659 \\
cassation sans renvoi & 2,078 \\
cassation partielle & 9,543 \\
cassation partielle sans renvoi & 1,015 \\
\emph{cassation partielle cassation} & 1,162 \\
\emph{cassation partielle rejet cassation} & 906 \\
rejet & 67,981 \\
irrecevabilite & 2,376 \\
\hline
\end{tabular}
\end{center}
\caption{\label{tab:2} Distribution of cases according to ruling type.}
\end{table}


\noindent Finally, in task 3, we investigate whether the text of the case description contained indicators of the period when it was written, a popular NLP task called temporal text classification addressed by a recent SemEval task \cite{popescu2015semeval}. Table ~\ref{tab:3} shows the distribution of cases in each decade. Due to the amount of cases, we grouped all cases dated 1959 and before in a single class. We run temporal text classification experiments with 7 classes. Table ~\ref{tab:3} shows the distribution of cases per decade.

\begin{table}[!ht]
\begin{center}
\begin{tabular}{lr}
\hline 
\bf Time Span & \bf \# of cases \\ \hline

Until 1959	& 201 \\
1960 - 1969 & 4,797  \\
1970 - 1979			& 23,964 \\
1980 - 1989	& 18,233 \\
1990 - 1999			& 16,693 	\\
2000 - 2009	& 12,577 \\
2010 - 2016		& 4,541\\

\hline
\end{tabular}
\end{center}
\caption{\label{tab:3} Distribution of cases in seven time intervals.}
\end{table}

\noindent For the three tasks we eliminated the occurrence of each word of the label from the text of the corresponding case description following the methodology described in \c{S}ulea et al. (2017). For task 1, law area prediction, we eliminated all words contained in the label. 

For predicting the ruling, we eliminated the ruling words themselves from all case descriptions. Aiming at a complete masking of the ruling, we additionally looked at the top 20 most important features of each class to investigate whether some of them could be directly linked to the target label. In this step, we realized that the label was present both in its nominal form (e.g. cassation, irrecevabilite) and in its verbal form (e.g. casse, casser) and eliminated both. For the task of predicting the century and decade in which a particular ruling took place, we eliminated all digits from the case description text, even though some of the digits referred to cited laws. 


\section{Methodology}

We approach the three tasks using a system based on classifier ensembles. Classifier ensembles have proven to achieve high performance in many task classification tasks such as grammatical error detection \cite{xiang2015chinese}, complex word identification \cite{malmasi2016LTG}, identifying self-harm risk in mental health forums \cite{malmasi:2016:clpsych}, and dialect identification \cite{dsl2016}.

There are many types of classifier ensembles and in this work we apply a mean probability classifier. The method works by adding probability estimates for each class together and assigning the class label with the highest average probability as the prediction. By using probability outputs in this way a classifier's support for the true class label is taken into account, even when it is not the predicted label (e.g. it could have the second highest probability).
This method is considered to be simple and it has been shown to work well on a wide range of problems. It is intuitive, stable \cite{kuncheva2014combining} and resilient to estimation errors \cite{kittler1998combining} making it one of the most robust combiners described in the literature.

As features, our system uses word unigrams and word bigrams. To evaluate the success of our method we compare the results obtained by the mean probability ensemble system with the results reported in \c{S}ulea et al. (2017) who approach the three tasks described in this paper using the scikit-learn implementation \cite{skl} of the LIBLINEAR SVM classifier \cite{liblinear} trained on bag of words and bag of bigrams.

Finally, as to the evaluation, we employ a stratified 10-fold cross-validation setup for all experiments. We chose this approach to be able to compare our results with those reported by \c{S}ulea et al. (2017) and also to take the inherent imbalance of the classes present in the dataset into account. We report results in terms of average precision, recall, F1 score, and accuracy for all classes. 

\section{Results}

\subsection{Law Area}
\label{sec:area}

In our first experiment, we trained our system to predict the law area of a case, given its case description preprocessed as described in Section 3 (i.e. removing all "give-away" references in the original data to simulate a realistic draft case description scenario, where the prediction - here in task 1 law area - is not already preempted). Table ~\ref{tab:res-la1} reports the average precision, recall, f1 score, and accuracy scores obtained of our method when discriminating between the aforementioned 8 classes each of them containing at least 200 instances. The scores reported by \c{S}ulea et al. (2017) using the same dataset are presented for comparison. 

\vspace{5mm}

\begin{center}
\begin{table}[!ht]
\begin{tabular}{lllll} 
\hline 
\bf Model 		& \bf P & \bf R 	& \bf F1 	& \bf Acc.\\\hline
Ensemble 			& 96.8\%		& 96.8\%	& \bf 96.5\%  & 96.8\% \\
\c{S}ulea et al. (2017)		& 90.9\%		& 90.2\%	& 90.3\% 	& 90.2\%\\
\hline
\end{tabular}
\caption{Classification results for law area prediction.}
\label{tab:res-la1}
\end{table}
\end{center}

\vspace{-2mm}

\begin{figure}[!ht]
\centering
\includegraphics[width=.49\textwidth]{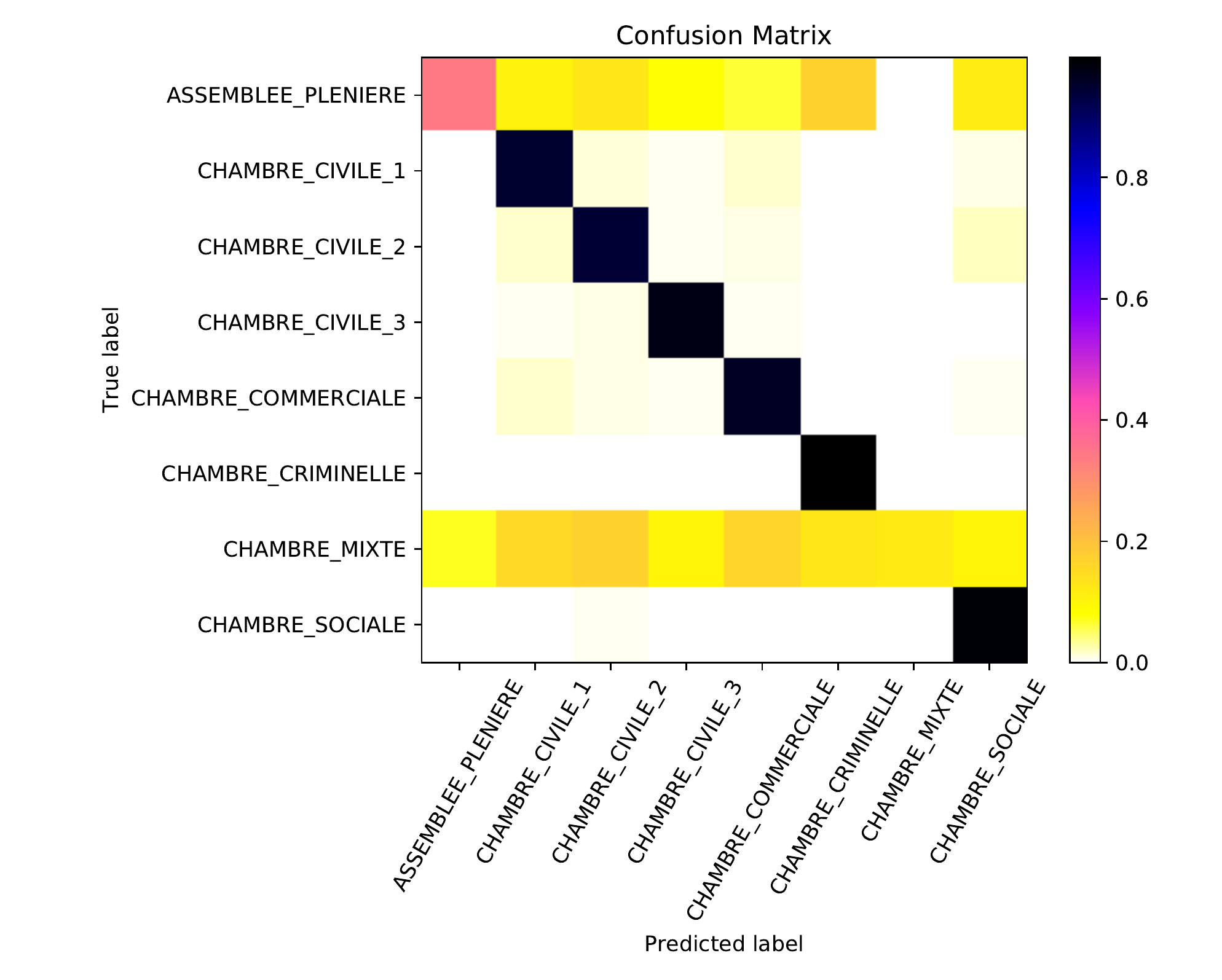}
\caption{Confusion matrix for law area prediction.}
\label{fig:area}
\end{figure}

\noindent We observe that the ensemble method outperforms the liner SVM classifier by a large margin, 96.8\% accuracy compared to 90.3\% reported by \c{S}ulea et al. (2017). We investigate the performance of the ensemble system for each individual class by looking at the confusion matrix presented in Figure \ref{fig:area}.

\vspace{2mm}

\vspace{2mm}

\noindent The confusion matrix presented in Figure \ref{fig:area} shows that cases from the {\em chambre mixte} are the most difficult predict. This is firstly because this class and {\em assemblee pleniere}, the second most difficult class to predict, contain the two lowest numbers of instances in the dataset (222 and 544 respectively), and secondly because by nature the {\em chambre mixte} received mixed cases from other courts such as civil and commercial.

\subsection{Case Ruling}
\label{sec:ruling}

The results for the second task, court ruling prediction, are presented in Table ~\ref{tab:res-rul}. We report the results obtained in both experiment setups, the 6-class setup and in the 8-class setup. The mean probability ensemble once again outperforms the method by \c{S}ulea et al. (2017) in both settings. We observe a 2.9 percentage point decrease in absolute average f1 score when the ensemble classifier is trained on the dataset with more classes which is explained by the increase in number of classes from 6 to 8 leading to a more challenging classification scenario. 

\vspace{5mm}

\begin{table}[!ht]
\centering
\begin{tabular}{clllll} 
\hline 
 \bf Classes & \bf Model & \bf P & \bf R & \bf F1 	& \bf Acc.\\\hline
6 & Ensemble	& 98.6\%	& 98.6\%	& \bf 98.6\%  	& 98.6\% \\
6 & \c{S}ulea et al. (2017)		& 97.1\%	& 96.9\%	& 97.0\% 	& 96.9\% \\
\hline
8 & Ensemble 	& 95.9\%	& 96.2\%	& \bf 95.8\%	& 96.2\% \\
8 & \c{S}ulea et al. (2017)		& 93.2\%	& 92.8\%	& 92.7\% 	& 92.8\% \\
\hline
\end{tabular}
\caption{Classification results for ruling prediction.}
\label{tab:res-rul}
\end{table}

\noindent To better understand the difficulties faced by our method in discriminating between the ruling classes we first looked at the list of the most informative unigrams for each class. We found a few clear cases of top-ranked words that are related to the target class, but even so the analysis did not go that far indicating that a more interesting analysis is only possible without the aid of an expert in French law. 

\begin{figure}[!ht]
\centering
\includegraphics[width=.49\textwidth]{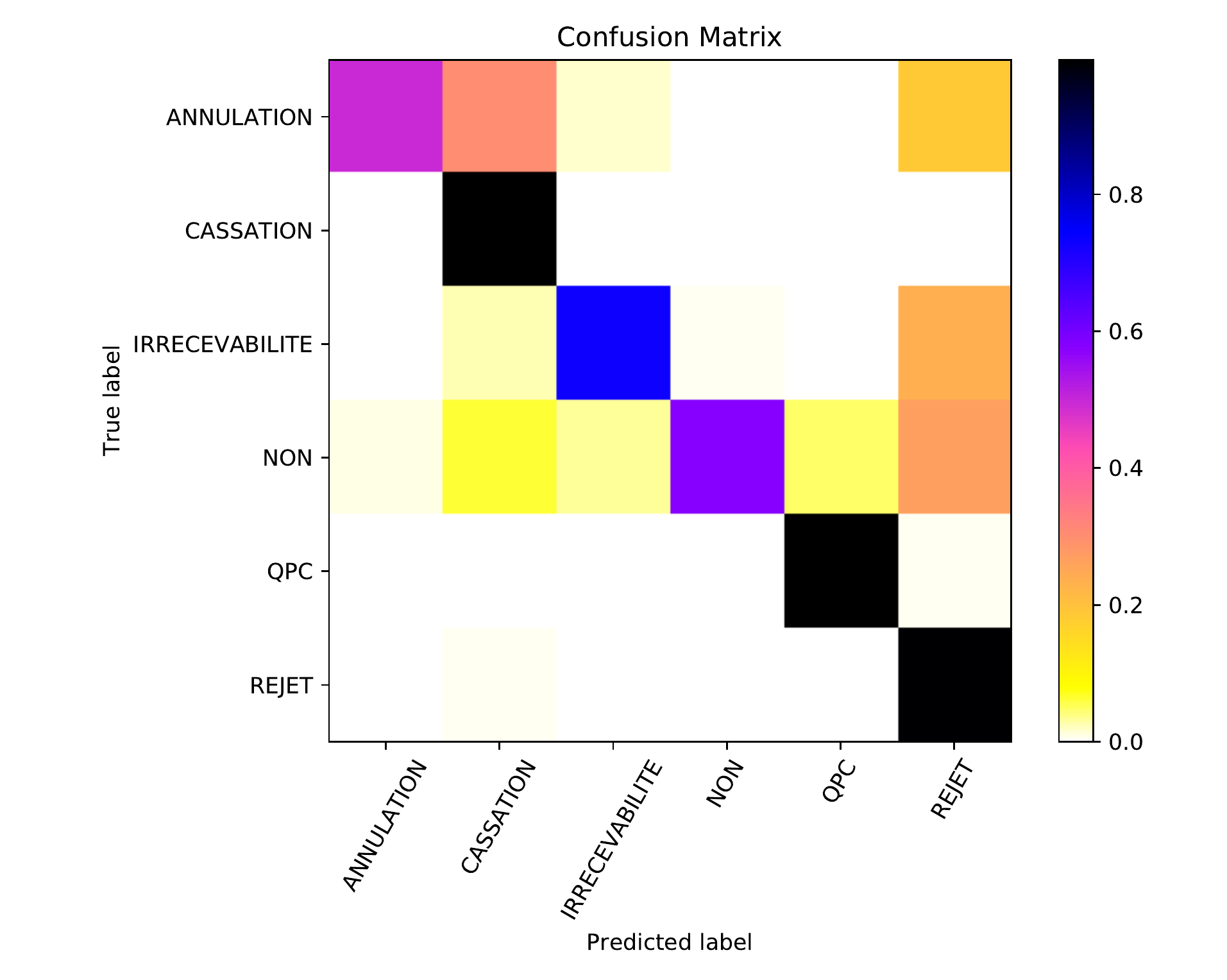}
\caption{Confusion matrix for law area prediction.}
\label{fig:ruling}
\end{figure}

Subsequently, we looked at the confusion matrix of predictions. In Figure \ref{fig:ruling} we present a confusion matrix of the performance obtained for each individual class in the 6-class setup experiment. We observe that the two most difficult classes for the system were {\em non-lieu} and {\em annulation}. These two classe are also the two classes which contained the least amount of examples which probably led to the poor performance of the classifier in identifying instances from these classes.


\vspace{3mm}

\subsection{Temporal Text Classification}
\label{sec:temporal}

Finally, in Table \ref{tab:res-date} we present the results obtained in the third set of experiments described in this paper, predicting the time span of cases and rulings in a 7-class setting. Again all data was preprocessed as indicated in Section 3.

\vspace{5mm}

\begin{table}[!ht]
\centering
\begin{tabular}{lllll} 
\hline  
\bf Model & \bf P & \bf R & \bf F1 	& \bf Acc. \\
\hline
Ensemble & 87.3\%	& 87.0\%	& \bf 87.0\% 	& 87.0\% \\
\c{S}ulea et al. (2017)	& 75.9\%	& 74.3\%	& 73.2\% & 74.3\% \\
 \hline
\end{tabular}
\caption{Classification results for temporal prediction.}
\label{tab:res-date}
\end{table}

\noindent Results obtained by the ensemble system in this experiment outperform the method by \c{S}ulea et al. (2017) by a large margin. This outcome once again confirms the robustness of classifier ensembles for many text classification tasks including those presented in this paper. The mean probability ensemble system achieved 87\% f1 score against 73.2\% reported by \c{S}ulea et al. (2017). 

The results obtained by our system in the temporal text classification task suggest that classifier ensembles are a good fit for predicting the publication date not only of legal texts but other types of texts as well. This is a particularly relevant application for researchers in the digital humanities who are often working with manuscripts with unknown or uncertain publication date. The use of ensembles for this task is, to the best of our knowledge, under explored and should be investigated further. 

It should be noted, however, that predictions in this experiment are only estimates as the definition of time spans in unities such as month, year, or decade (in the case of this paper) is arbitrary. Previous work in temporal text classification stressed that supervised methods, such as the one presented in this paper fail to capture the linearity of time \cite{niculae2014temporal,zampieri2016lrec}. Other methods, such as ranking or regression, could be applied to obtain more accurate predictions.

\section{Conclusions and Future Work}

In this paper we investigated the application of text classification methods to the legal domain using the cases and rulings of the French Supreme Court. We showed that a system based on SVM ensembles can obtain high scores in predicting the law area and the ruling of a case, given the case description, and the time span of cases and rulings. The ensemble method presented in this paper outperformed a previously proposed \c{S}ulea et al. (2017) using the same dataset. 

We applied computational methods to mask the case description attached to a judge's ruling so that they convey as little information as possible about the ruling. This simulates the knowledge a lawyer would have prior to entering court. 

The work presented in this paper confirms that text classification techniques can indeed be used to provide valuable assistive technology base as support for law professionals in obtaining guidance and orientation from large corpora of previous court rulings. In future work, we would like to investigate the extent to which a more accurate draft form can be induced from the court's case description.



\section*{Acknowledgements}

Parts of this work have been carried out while the first and the second author, Octavia-Maria \c{S}ulea and Marcos Zampieri, were at the German Research Center for Artificial Intelligence (DFKI).

We would like to thank the anonymous reviewers for providing us with constructive feedback and suggestions.

\bibliographystyle{ACM-Reference-Format}
\bibliography{bibliography}

\end{document}